\documentclass[final]{cvpr}

\usepackage{times}
\usepackage{epsfig}
\usepackage{graphicx}
\usepackage{amsmath}
\usepackage{amssymb}
\usepackage{multirow}

\usepackage{color}

\usepackage[pagebackref=true,breaklinks=true,colorlinks,bookmarks=false]{hyperref}

\newcommand{\tablestyle}[2]{\setlength{\tabcolsep}{#1}\renewcommand{\arraystretch}{#2}\centering\small}
\newlength\savewidth\newcommand\shline{\noalign{\global\savewidth\arrayrulewidth
  \global\arrayrulewidth 1pt}\hline\noalign{\global\arrayrulewidth\savewidth}}

\def\cvprPaperID{****} 
\def\confYear{CVPR 2021}

\begin{document}

\title{Towards Training Stronger Video Vision Transformers 
\\
for EPIC-KITCHENS-100 Action Recognition}

\author{
Ziyuan Huang$^{1,4\dag}$
\quad Zhiwu Qing$^{2,4\dag}$ 
\quad Xiang Wang$^{2,4}$ 
\quad Yutong Feng$^{3,4}$ 
\quad Shiwei Zhang$^{4*}$
\\
\quad Jianwen Jiang$^4$ 
\quad Zhurong Xia$^4$ 
\quad Mingqian Tang$^4$ 
\quad Nong Sang$^{2}$ 
\quad Marcelo H. Ang Jr$^{1*}$
\\
$^1$ARC, National University of Singapore\\
$^2$ 
AIA, Huazhong University of Science and Technology\\
$^3$Tsinghua University \quad
$^4$Alibaba Group\\
{\tt\small ziyuan.huang@u.nus.edu, mpeangh@nus.edu.sg}\\
{\tt\small \{qzw, wxiang, nsang\}@hust.edu.cn}\\
{\tt\small fyt19@mails.tsinghua.edu.cn}\\
{\tt\small \{zhangjin.zsw, jianwen.jjw, zhurong.xzr, mingqian.tmq\}@alibaba-inc.com}
}

\maketitle

\let\thefootnote\relax\footnotetext{$\dag$ Equal Contribution.}
\let\thefootnote\relax\footnotetext{$*$ Corresponding authors.}
\let\thefootnote\relax\footnotetext{This work is done when X. Wang, Z. Qing, Z. Huang and Y. Feng are interns at Alibaba Group.}

\begin{abstract}
   With the recent surge in the research of vision transformers, they have demonstrated remarkable potential for various challenging computer vision applications, such as image recognition, point cloud classification as well as video understanding. 
   In this paper, we present empirical results for training a stronger video vision transformer on the EPIC-KITCHENS-100 Action Recognition dataset. 
   Specifically, we explore training techniques for video vision transformers, such as augmentations, resolutions as well as initialization, etc. 
   With our training recipe, a single ViViT model achieves the performance of 47.4\% on the validation set of EPIC-KITCHENS-100 dataset, outperforming what is reported in the original paper~\cite{arnab2021vivit} by 3.4\%.
   We found that video transformers are especially good at predicting the noun in the verb-noun action prediction task. 
   This makes the overall action prediction accuracy of video transformers notably higher than convolutional ones. 
   Surprisingly, even the best video transformers underperform the convolutional networks on the verb prediction. 
   Therefore, we combine the video vision transformers and some of the convolutional video networks and present our solution to the EPIC-KITCHENS-100 Action Recognition competition.
\end{abstract}

\section{Introduction}
Recent developments in the computer vision field have witnessed rapid expansion of transformer based model family, which has demonstrated remarkable potential in various computer vision applications, such as image recognition~\cite{dosovitskiy2020vit,zhou2021deepvit}, point cloud classification~\cite{zhao2020pointtransformer} as well as video understanding~\cite{arnab2021vivit,bertasius2021timesformer}. They are shown to supersede the performance of convolutional networks when given proper combinations of augmentation strategies~\cite{touvron2020deit}. 

In this paper, we report our recent exploration on the training techniques for the video vision transformers. Specifically, we employ ViViT~\cite{arnab2021vivit} as our base model, and explored the influence of the quality of the data source, augmentations, input resolutions as well as the initialization of the network. The resultant training techniques enable ViViT to achieve $47.4\%$ on the action recognition accuracy of Epic-Kitchen-100 dataset. Additionally, it is noticed that although ViViT performs better than convolutional networks by a notable margin on the action classification, it underperforms convolutional ones on verb classification. This means that the ensemble of them can be beneficial to increasing the final accuracy. By combining video transformers with the convolutional ones, this paper finally presents our solution to the Epic-Kitchen-100 Action Recognition challenge. 

\section{Training video vision transformers}

We use the ViViT-B/16x2 with factorized encoder as our base model. Two classification heads are connected to the same class token to predict the verb and the noun for the input video clip respectively.
We first pre-train the networks on large video datasets that are available publicly, and then fine-tune the ViViT on the epic-kitchen dataset. 

\subsection{Initialization preparation}
There are multiple ways to prepare the pre-trained model, such as supervised pre-training~\cite{tran2019csn,feichtenhofer2019slowfast,arnab2021vivit,carreira2017i3d} as is used in~\cite{song2019tacnet,qing2021tca,wang2020cbr} as well as unsupervised ones~\cite{huang2021mosi,han2020coclr}.
Here we adopt the supervised pre-training as it yields a better downstream performance.
The model is firstly trained on Kinetics 400~\cite{kay2017k400}, Kinetics 700~\cite{carreira2019k700} and Something-Something-V2~\cite{goyal2017ssv2}. respectively. For this part, we mostly follow the training recipe in DeepViT~\cite{zhou2021deepvit} to boost the performance. Specifically, we use AdamW as our optimizer~\cite{loshchilov2017adamw} and set the base learning rate to 0.0001. The weight decay is set to 0.1. We initialize the ViViT model with the ViT weight pre-trained on ImageNet21k following the initialization approach in \cite{arnab2021vivit}, and train the model for 30 epochs with cosine learning rate schedule. The training is warmed up with 2.5 epochs, with the start learning rate as 1e-6. We enable color augmentation, mixup and label smoothing. The model is additionally regularized with a droppath rate of 0.2. The results on the Kinetics and SSV2 are as in Table~\ref{tab:vivit-k400-k700-pt}. We also trained ViViT on the optical flow of Kinetics 400, which we extract using Raft~\cite{teed2020raft} and TVL1~\cite{zach2007tvl1}.

\begin{table}[t]
    \centering
\tablestyle{4pt}{1.0}
\small
\begin{tabular}{c|cc|c|c}
    \normalsize Model & \normalsize Dataset & \normalsize Resolution & \normalsize Top 1 & \normalsize Views\\
    \shline
    ~ & \multirow{2}{*}{K400} & 224 & 78.6 & $4\times3$ \\
    ~ & ~ & 320 & 80.6 & $4\times3$ \\
    \cline{2-5}
    ~ & \multirow{2}{*}{K700} & 224 & 69.7 & $4\times3$  \\
    ViViT-B/16x2 & ~ & 320 & 71.5 & $4\times3$ \\
    \cline{2-5}
    Fact. encoder & \multirow{2}{*}{SSV2} & 224 & 63.6 & $1\times1$\\
    ~ & ~ & 320 & - & $1\times1$\\
    \cline{2-5}
    ~ & K400-Raft & 224 & 60.5 & $4\times3$ \\ 
    ~ & K400-Tvl1 & 224 & 65.4 & $4\times3$ \\ 
\end{tabular}\\
    \caption{\textbf{Pre-training ViViT on Kinetics 400, 700 and SSV2.} The pre-training using respective dataset X with an input resolution Y is denoted further as X-Y. \textit{E.g.,} K400-224 for initialization weights trained on K400 with an input resolution of 224. }
    \label{tab:vivit-k400-k700-pt}
\end{table}

\begin{table}[t]
    \centering
\tablestyle{4pt}{1.0}
\small
\begin{tabular}{c|c|ccc|ccc|c}
    \multirow{2}{*}{\normalsize ID} & \multirow{2}{*}{\normalsize Init.} & \multirow{2}{*}{\normalsize Qual.} & \multirow{2}{*}{\normalsize Res.} & \multirow{2}{*}{\normalsize Aug.} & \multicolumn{4}{c}{\normalsize Top1} \\
    ~ & ~ & ~ & ~ & ~ & A & V & N & A* \\
    \shline
    A & IN21K & 256 & 224 & CJ & 36.1 & 62.4 & 48.2 & -  \\ 
    B & \textcolor{blue}{\textbf{K400-224}} & 256 & 224 & CJ & 37.2 & 61.7 & 50.9 & - \\ 
    C & K400-224 & \textcolor{blue}{\textbf{512}} & 224 & CJ & 38.4 & 62.7 & 52.2 & - \\ 
    D & \textcolor{blue}{\textbf{K700-224}} & 512 & 224 & CJ & 39.6 & 63.5 & 53.3 & - \\ 
    E & K700-224 & 512 & 224 & \textcolor{blue}{\textbf{CJ+}} & 42.8 & 65.2 & 56.2 & - \\ 
    \hline
    \multirow{2}{*}{F} & \multirow{2}{*}{K700-224} & \multirow{2}{*}{512} & \multirow{2}{*}{\textcolor{blue}{\textbf{320}}} & \multirow{2}{*}{CJ+} & 45.2 & \multirow{2}{*}{67.4} & \multirow{2}{*}{58.9} & 42.4\\ 
    ~ & ~ & ~ & ~ & ~ & 46.3\dag & ~ & ~ & 43.4\\
    \hline
    \multirow{2}{*}{G} & \multirow{2}{*}{\textcolor{blue}{\textbf{SSV2-224}}} & \multirow{2}{*}{512} & \multirow{2}{*}{320} & \multirow{2}{*}{CJ+} & 44.5 & \multirow{2}{*}{\textbf{67.5}} & \multirow{2}{*}{57.5} & -\\ 
    ~ & ~ & ~ & ~ & ~ & 45.7\dag & ~ & ~ & -\\
    \hline
    \multirow{2}{*}{H} & \multirow{2}{*}{K700-224} & \multirow{2}{*}{512} & \multirow{2}{*}{\textcolor{blue}{\textbf{384}}} & \multirow{2}{*}{CJ+} & 45.8 & \multirow{2}{*}{67.2} & \multirow{2}{*}{\textbf{59.0}} & 42.5\\ 
    ~ & ~ & ~ & ~ & ~ & \textbf{47.0}\dag & ~ & ~ & -\\
    \hline
    \cite{arnab2021vivit} & - & - & 224 & CJ* & 44.0 & 66.4 & 56.8 & -
    
\end{tabular}\\
    \caption{\textbf{Fine-tuning ViViT on EPIC-KITCHENS-100.} 
    \textit{Init} indicates the pre-training dataset. 
    \textit{Qual} indicates the length of the short side of the input video before any transformation is performed. We resize the original video. 
    \textit{Res} indicates the resolution of the input video to the model. 
    \textit{Aug} indicates the augmentation strategy besides random cropping and random flipping. 
    \textit{A, V} and \textit{N} denotes respectively the action, verb and noun prediction accuracies.
    \textit{A*} denotes the action prediction accuracy on the test set.
    \textit{CJ+} and \textit{CJ} respectively denote random color jitter with and without mixup, cutmix with random erasing.
    \textit{CJ*} indicates a different augmentation strategy used in \cite{arnab2021vivit}.
    \dag indicates that the action prediction is calculated for each view first before aggregating them together.
    \textcolor{blue}{\textbf{Blue font}} highlights the change in the respective experiment. 
    \textbf{Bold font} in the performance columns indicate the best performing model. }
    \vspace{-3mm}
    \label{tab:vivit-ek100}
\end{table}

\subsection{Training video transformers on Epic-Kitchen}
For training video transformers on Epic-Kitchen, we ablate on the training recipes in terms of initialization, the quality of data source, augmentations, input resolutions, the action calculation strategy, as well as the temporal sampling stride. The training parameters including the optimizer, the base learning rate. The training schedule is set to be overall 50 epochs and warmup for 10 epochs. The results can be observed in Table~\ref{tab:vivit-ek100}. If not otherwise specified, we sample frames with one as the interval.

\noindent\textbf{Initialization}: we ablate initialization by pre-training from ImageNet-21K, Kinetics400, Kinetics700 as well as SSV2. The reason that we also ablated the SSV2 initialization is that SSV2 is also egocentric action recognition datasets with complex spatio-temporal interactions. It can be observed that using a strong initialization (from ImageNet21K to Kinetics400, and further to Kinetics700) lead to a notable improvement on the action recognition accuracy. If we decompose the improvement on verb and noun predictions, we can see that stronger initialization model brings the most improvements on noun predictions. However, although higher verb prediction accuracy can be observed by replacing the initialization from K700 to SSV2 (0.1\%), there is a notable decrease on the noun prediction (1.4\%). Therefore, in the final submission, we did not include models initialized with SSV2.

\noindent\textbf{Quality of data source}: to mitigate the pressure on the hard drive i/o and thus to speed up training, we resize the short side of the videos to 256 and 512 respectively. It can be observed that raising the quality of the input data source can have an improvement of 1.2\%, 1.0\% and 1.3\% on the action, verb and noun predictions respectively. 

\noindent\textbf{Augmentations}: we observe the benefit of utilizing stronger augmentations (mixup~\cite{zhang2017mixup}, cutmix~\cite{yun2019cutmix} and random erasing~\cite{zhong2020randomerasing}). Compared to only using random color jittering, stronger augmentations brings an improvement of 3.2\% on the action prediction, and 1.7\% as well as 2.9\% on the verb and noun predictions respectively. 

\noindent\textbf{Input resolutions}: we further alter the input resolutions. Raising input resolution from 224 to 320 brings about 2.4\% improvement on the action prediction, 2.2\% on verb prediction and 2.7\% on the noun prediction. A saturation of the prediction accuracy is observed when we further raise the input resolution from 320 to 384, where only an improvement of 0.6\% on the action prediction is observed. 

\noindent\textbf{Action score calculation}: as indicated in the table as numbers with \dag, calculating action scores differently could lead to different action prediction results. As two predictions are made for each video clip, there are two ways of aggregating action predictions for multiple views. Suppose we have predictions for verb $P_v^i\in\mathbb{R}^{1\times N_v}$ and noun $P_n^i\in\mathbb{R}^{1\times N_n}$ respectively, where $N_v$ and $N_n$ denotes the number of class for verbs and nouns and $i$ denotes the index for a view, the first way of aggregating the predictions are:

\begin{equation}
    P_a = (\sum_i P_v^\top)  (\sum_i P_n) \ ,
\end{equation}
\noindent where $P_a\in \mathbb{R}^{N_v\times N_n}$ denotes the prediction for actions.
This approach aggregates the verb and noun predictions for multiple views first, before calculating the action predictions directly. The second approach calculate the action prediction for each view respectively, before aggregating them:

\begin{equation}
    P_a = \sum_i (P_v^\top  P_n) \ .
\end{equation}

As can be seen from the Table~\ref{tab:vivit-ek100}, aggregating action scores for each view can outperform the other variant by around 1\%. What's more important is that this improvement can be reflected in the test set as well. 

\begin{table}[t]
    \centering
\tablestyle{4pt}{1.0}
\small
\begin{tabular}{c|c|ccc}

    \multirow{2}{*}{\normalsize ID} & \normalsize Temp  & \multicolumn{3}{c}{\normalsize Top1} \\
    ~ & \normalsize Sampling Rate & A & V & N \\
    \shline
    \multirow{2}{*}{F} & \multirow{2}{*}{\normalsize 2} & 45.2 & \multirow{2}{*}{67.4} & \multirow{2}{*}{58.9}\\
    ~ & ~ & 46.3\dag & ~ & ~ \\
    \hline
    \multirow{2}{*}{I} & \multirow{2}{*}{\normalsize 3} & 46.4 & \multirow{2}{*}{68.4} & \multirow{2}{*}{59.6}\\
    ~ & ~ & 47.4\dag & ~ & ~ \\
    \hline
    \cite{arnab2021vivit} & 2 & 44.0 & 66.4 & 56.8

\end{tabular}\\
    \caption{\textbf{Altering the temporal sampling rate of ViViT.}
    \textit{A, V} and \textit{N} denotes respectively the action, verb and noun prediction accuracies.
    \dag indicates that the action prediction is calculated for each view first before aggregating them together.
    }
    \label{tab:stride}
\end{table}

\begin{table}[t]
    \centering
\tablestyle{4pt}{1.0}
\small
\begin{tabular}{c|c|ccc}

    \multirow{2}{*}{\normalsize Model} & \normalsize Optical  & \multicolumn{3}{c}{\normalsize Top1} \\
     ~ & \normalsize Flow & A & V & N \\
    \shline
    \multirow{2}{*}{ViViT-B/16x2-Flow-A} & \multirow{2}{*}{Raft} & 34.6 & \multirow{2}{*}{66.8} & \multirow{2}{*}{43.5}\\
    ~ & ~ & 35.4\dag & ~ & ~ \\
    \hline
    \multirow{2}{*}{ViViT-B/16x2-Flow-B} & \multirow{2}{*}{TVL1} & 34.5 & \multirow{2}{*}{66.4} & \multirow{2}{*}{43.3}\\
    ~ & ~ & 35.1\dag & ~ & ~ \\

\end{tabular}\\
    \caption{\textbf{Training ViViT with optical flow.}
    \textit{A, V} and \textit{N} denotes respectively the action, verb and noun prediction accuracies.
    \dag indicates that the action prediction is calculated for each view first before aggregating them together.}
    \label{tab:stride}
\end{table}

\begin{table}[t]
    \centering
\tablestyle{4pt}{1.0}
\small
\begin{tabular}{c|c|ccc}
    \multirow{2}{*}{\normalsize Model} & \multirow{2}{*}{\normalsize Training}  & \multicolumn{3}{c}{\normalsize Top1} \\
    ~ & ~ & A & V & N \\
    \shline
    \multirow{3}{*}{TimeSformer $8\times32$} & original & 34.4 & 57.1 & 51.3\\
    ~ & ours-224 & 39.4 & 63.9 & 51.7 \\
    ~ & ours-320 & \textbf{42.5} & \textbf{65.2} & \textbf{55.0}
    
\end{tabular}\\
    \caption{\textbf{Results of TimeSformer on EPIC-KITCHENS-100.}
    \textit{A, V} and \textit{N} denotes respectively the action, verb and noun prediction accuracies.
    All action accuracies are obtained by calculating action predictions before aggregating them together.
    }
    \label{tab:stride}
\end{table}

\noindent\textbf{Temporal sampling stride.} Since Epic-Kitchen dataset has a relatively higher FPS, sampling frames with one frame as the interval (which means the temporal sampling rate is 2) can be insufficient for the temporal coverage. When sampling 32 frames, only one second is covered. Therefore, we also ablated on the temporal sampling rate, and the result is presented in Table~\ref{tab:stride}. As can be seen, a minor modification on the temporal sampling rate can have notable improvement on the performance. One reason for this may be the longer temporal coverage. Another possible reason is that the resultant FPS genearted by the sampling rate of 3 is closer to the pretraining FPS. Our final single model performance of ViViT-B/16x2-I outperforms the reported performance in \cite{arnab2021vivit} by 3.4\%.

\subsection{Training video transformers with optical flow}
In order to capture better motion features, optical flow is utilized as another data source. The video transformers with optical flow as the data source are trained using the same training recipe as mentioned before. We trained two optical flow models, whose inputs are respectively optical flow extracted using Raft~\cite{teed2020raft} and TVL1~\cite{zach2007tvl1}. The results are presented in

\subsection{Other transformer based models}
Another transformer based video classification model that we use is the TimeSformer~\cite{bertasius2021timesformer}. For TimeSformer, we directly use the open-sourced pretrained model on K600 and first kept its default settings and trained for 15 epochs. Then we used our training recipe in comparison. It shows that our training recipe improves the original one by 5\% on the action prediction accuracy. Further increasing the input resolution gives an improvement of 3\% on the action prediction accuracy.

\begin{table}[t]
    \centering
\tablestyle{1.5pt}{1.0}
\small
\begin{tabular}{c|c|ccc|ccc|c}
    
    \multirow{2}{*}{\normalsize ID} & \multirow{2}{*}{\normalsize Model} & \multirow{2}{*}{\normalsize F. BN} & \multirow{2}{*}{\normalsize Res.} & \multirow{2}{*}{\normalsize Aug.} & \multicolumn{4}{c}{\normalsize Top1} \\
    ~ & ~ & ~ & ~ & ~ & A & V & N & A* \\
    \shline
    \multirow{2}{*}{A} & \multirow{2}{*}{ir-CSN-152} & \multirow{2}{*}{$\times$} & \multirow{2}{*}{224} & \multirow{2}{*}{CJ} & 41.0 & \multirow{2}{*}{66.4} & \multirow{2}{*}{52.4} & 37.8 \\
    ~ & ~ & ~ & ~ & ~ & 42.4\dag & ~ & ~ & -\\
    \hline
    \multirow{2}{*}{B} & \multirow{2}{*}{ir-CSN-152} & \textcolor{blue}{\multirow{2}{*}{$\checkmark$}} & \multirow{2}{*}{224} & \multirow{2}{*}{CJ} & 42.7 & \multirow{2}{*}{67.6} & \multirow{2}{*}{55.1} & - \\
    ~ & ~ & ~ & ~ & ~ & 43.9\dag & ~ & ~ & 40.9 \\
    \hline
    \multirow{2}{*}{C} & \multirow{2}{*}{ir-CSN-152} & \multirow{2}{*}{$\checkmark$} & \multirow{2}{*}{224} & \textcolor{blue}{\multirow{2}{*}{\textbf{CJ+}}} & 43.5 & \multirow{2}{*}{68.4} & \multirow{2}{*}{55.9} & - \\
    ~ & ~ & ~ & ~ & ~ & 44.5\dag & ~ & ~ & \textbf{42.5} \\
    \hline
    \multirow{2}{*}{D} & \multirow{2}{*}{ir-CSN-152} & \multirow{2}{*}{$\checkmark$} & \textcolor{blue}{\multirow{2}{*}{\textbf{320}}} & \multirow{2}{*}{CJ+} & 45.1 & \multirow{2}{*}{\textbf{69.0}} & \multirow{2}{*}{\textbf{57.2}} & - \\
    ~ & ~ & ~ & ~ & ~ & \textbf{46.2}\dag & ~ & ~ & 42.4 \\
    \shline
    \multirow{2}{*}{-} & \multirow{2}{*}{SlowFast-16$\times$8-101} & \multirow{2}{*}{$\checkmark$} & \multirow{2}{*}{224} & \multirow{2}{*}{CJ+} & 43.0 & \multirow{2}{*}{68.2} & \multirow{2}{*}{55.1} & -\\
    ~ & ~ & ~ & ~ & ~ & 43.9\dag & ~ & ~ & - \\
\end{tabular}\\
    \caption{\textbf{Fine-tuning ir-CSN-152 and SlowFast-16$\times$8-101 on EPIC-KITCHENS-100.} \textit{F.BN} denotes frozen batch norm mean and variance. 
    \textit{Res} indicates the resolution of the input video to the model. 
    \textit{Aug} indicates the augmentation strategy besides random cropping and random flipping. 
    \textit{A, V} and \textit{N} denotes respectively the action, verb and noun prediction accuracies.
    \textit{A*} denotes the action prediction accuracy on the test set.
    \dag indicates that the action prediction is calculated for each view first before aggregating them together.
    \textcolor{blue}{\textbf{Blue font}} highlights the change in the respective experiment. 
    \textbf{Bold font} in the performance columns indicate the best performing model.
    }
    \vspace{-4mm}
    \label{tab:cnn-ek100}
\end{table}

\section{Training convolutional video networks}
Although video vision transformers can have a strong performance, complementary predicions are also needed from the convolutional networks. As we will shown in the following parts, the convolutional networks such as CSN~\cite{tran2019csn} and SlowFast~\cite{feichtenhofer2019slowfast} are relatively stronger at predicting verbs.

We use the ir-CSN-152 and SlowFast-$16\times8$-101 as our base model. Similar to the training process in ViViT model, we obtain the pre-trained weights by training these two models on Kinetics 700~\cite{carreira2019k700}. 
For training on the EPIC-KITCHENS-100 dataset, we use the same training parameters as the ViViT, including the optimizer and the learning rate schedule, etc. We follow~\cite{damen2020ek100} and freeze the batch norm mean and variance during training. The results can be seen in Table~\ref{tab:cnn-ek100}. As can be seen, freezing batch norm mean and variance gives about $1.3\%$ improvement on the action recognition accuracy. Applying mixup, cutmix as well as random erasing yields further improvements both on the validation and the test set. However, different from the experimental result in ViViT, although increasing the input resolution indeed increases the performance on the validation set, the accuracy on the test set is not improved. Therefore, we keep the training resolution as 224 for SlowFast-16$\times$8-101 as well. It is interesting to see that the convolutional models can outperform most ViViT in terms of verb prediction even when the input resolution is only 224$\times$224.

In order to cover a longer period for one video clip, we additionally employ the long-term feature banks (LFB)~\cite{wu2019lfb} for the CSN models. For these experiments, we initialize the model with Epic-Kitchen trained ir-CSN-152s, and further train the model for 10 epochs with the same base learning rate as before, with 2 warm-up epochs. The results are shown in Table~\ref{tab:lfb}. When using the features extracted by the original model that is used for initializing the training, we observe an improvement for ir-CSN-152-C on the noun prediction. When using the ViViT feature as the feature bank, the noun predictions are further improved, thus notably improving the final action prediction accuracy. In comparison, the verb accuracy is hardly affected. 

\begin{table}[t]
    \centering
\tablestyle{2.5pt}{1.0}
\small
\begin{tabular}{c|c|c|ccc}
    
    \multirow{2}{*}{ID} & \multirow{2}{*}{\normalsize Model} & \multirow{2}{*}{\normalsize LFB Feature} & \multicolumn{3}{c}{\normalsize Top1} \\
    ~ & ~ & ~ & A & V & N  \\
    \shline
    - & ir-CSN-152-B & - & 43.9\dag & 67.6 & 55.1 \\
    E & ir-CSN-152-B & ir-CSN-152-B & 42.9\dag & 66.9 & 54.7\\
    F & ir-CSN-152-B & ViViT-B/16x2-F& 47.3\dag & 67.6 & 60.1\\
    \hline
    - & ir-CSN-152-C & - & 44.5\dag & 68.4 & 55.9 \\
    G & ir-CSN-152-C & ir-CSN-152-C & 44.8\dag & 68.1 & 56.8\\
    H & ir-CSN-152-C & ViViT-B/16x2-F & \textbf{47.3}\dag & \textbf{68.1} & \textbf{60.3}\\
\end{tabular}\\
    \caption{\textbf{Applying long-term feature banks for ir-CSN-152.} 
    \dag indicates that the action prediction is calculated for each view first before aggregating them together.
    \textbf{Bold font} in the performance columns indicate the best performing model.
    }
    \vspace{-2mm}
    \label{tab:lfb}
\end{table}

\section{Ensembling models}
To utilize the complementary predictions of different models, we ensembled a selected subset of the presented models. The selected model set is presented in Table~\ref{tab:ensemble}. The ensemble of models boost the performance of the best performing one by 4.3\% on the action prediction. The final test accuracy that we obtained is 48.5\% on action prediction, 69.2\% on verb prediction and 60.3\% on noun prediction.

\begin{table}[t]
    \centering
\tablestyle{7pt}{1.0}
\begin{tabular}{c|ccc}
    
    \multirow{2}{*}{Model Name} & \multicolumn{3}{c}{\normalsize Top1} \\
    ~  & A & V & N  \\
    \shline
    ir-CSN-152-B & 43.9 & 67.6 & 55.1\\
    ir-CSN-152-C & 44.5 & 68.4 & 55.9\\
    ir-CSN-152-F & 47.2 & 67.6 & 60.1\\
    ir-CSN-152-G & 44.8 & 68.1 & 56.8\\
    ir-CSN-152-H & 47.3 & 68.1 & 60.3\\
    SlowFast-16$\times$8-101 & 43.9 & 68.2 & 55.1\\
    ViViT-B/16x2-Flow-A & 35.4 & 66.8 & 43.5 \\
    ViViT-B/16x2-Flow-B & 35.1 & 66.4 & 43.3 \\
    ViViT-B/16x2-F & 46.3 & 67.4 & 58.9\\
    ViViT-B/16x2-H & 47.0 & 67.2 & 59.0\\
    ViViT-B/16x2-I & 47.4 & 68.4 & 59.6\\
    TimeSformer-320 & 42.5 & 65.2 & 55.0\\
    \shline
    Overall (Val) & 51.7 & 72.4 & 62.6\\
    Overall (Test) & 48.5 & 69.2 & 60.3\\
    
\end{tabular}\\
    \caption{\textbf{List of ensembled models.}
    All the performance listed in the table are to calculate action for each view before aggregation.
    }
    \vspace{-2mm}
    \label{tab:ensemble}
\end{table}

\section{Conclusion}
This paper presents our solution for the EPIC-KITCHENS-100 action recognition challenge. We set out to train a stronger video vision transformer, and reinforce its performance by ensembling multiple video vision transformers as well as convolutional video recognition models.

\noindent\textbf{Acknowledgement. }
This work is supported by the Agency for Science, Technology and Research (A*STAR) under its AME Programmatic Funding Scheme (Project A18A2b0046), the National Natural Science Foundation of China under grant 61871435 and the Fundamental Research Funds for the Central Universities no. 2019kfyXKJC024 and by Alibaba Group through Alibaba Research Intern Program.

{\small
\bibliographystyle{ieee_fullname}
\bibliography{egbib}
}

\end{document}